\def\year{2022}\relax
\documentclass[letterpaper]{article} 
\usepackage{aaai22}  
\usepackage{times}  
\usepackage{helvet}  
\usepackage{courier}  
\usepackage[hyphens]{url}  
\usepackage{graphicx} 
\usepackage{natbib}  
\usepackage{caption} 
\DeclareCaptionStyle{ruled}{labelfont=normalfont,labelsep=colon,strut=off} 
\usepackage{algorithm}
\usepackage{algorithmic}
\usepackage{booktabs}
\usepackage[table,xcdraw]{xcolor}
\usepackage{subfigure}
\usepackage{multirow}
\usepackage{amsmath}
\usepackage{enumitem}
\usepackage{bbding}
\usepackage{float}
\usepackage{amsfonts}

\usepackage{newfloat}
\usepackage{listings}
\floatstyle{ruled}
\newfloat{listing}{tb}{lst}{}
\floatname{listing}{Listing}

\newcommand\correspondingauthor{\thanks{Corresponding author.}}
\title{Guide Local Feature Matching by Overlap Estimation}
\author {
    Ying Chen\textsuperscript{\rm 1}\equalcontrib,
    Dihe Huang\textsuperscript{\rm 2}\equalcontrib,
    Shang Xu\textsuperscript{\rm 1},
    Jianlin Liu\textsuperscript{\rm 1},
    Yong Liu\textsuperscript{\rm 1}\correspondingauthor
}
\affiliations {
    \textsuperscript{\rm 1}Tencent Youtu Lab\\
    \textsuperscript{\rm 2}Tsinghua University\\
    \{mumuychen, shangxu, jenningsliu, choasliu\}@tencent.com,
    hdh20@mails.tsinghua.edu.cn
}

\begin{document}

\maketitle

\begin{abstract}
Local image feature matching under large appearance, viewpoint, and distance changes is challenging yet important. Conventional methods detect and match tentative local features across the whole images, with heuristic consistency checks to guarantee reliable matches. In this paper, we introduce a novel Overlap Estimation method conditioned on image pairs with TRansformer, named OETR, to constrain local feature matching in the commonly visible region. OETR performs overlap estimation in a two-step process of feature correlation and then overlap regression. As a preprocessing module, OETR can be plugged into any existing local feature detection and matching pipeline, to mitigate potential view angle or scale variance. Intensive experiments show that OETR can boost state-of-the-art local feature matching performance substantially, especially for image pairs with small shared regions. The code will be publicly available at https://github.com/AbyssGaze/OETR.
\end{abstract}

\begin{figure}[t] 
\centering
\subfigure[Local feature matching by SuperPoint and SuperGlue.]{
	\begin{minipage}[b]{0.45\textwidth}
		\includegraphics[width=1\textwidth]{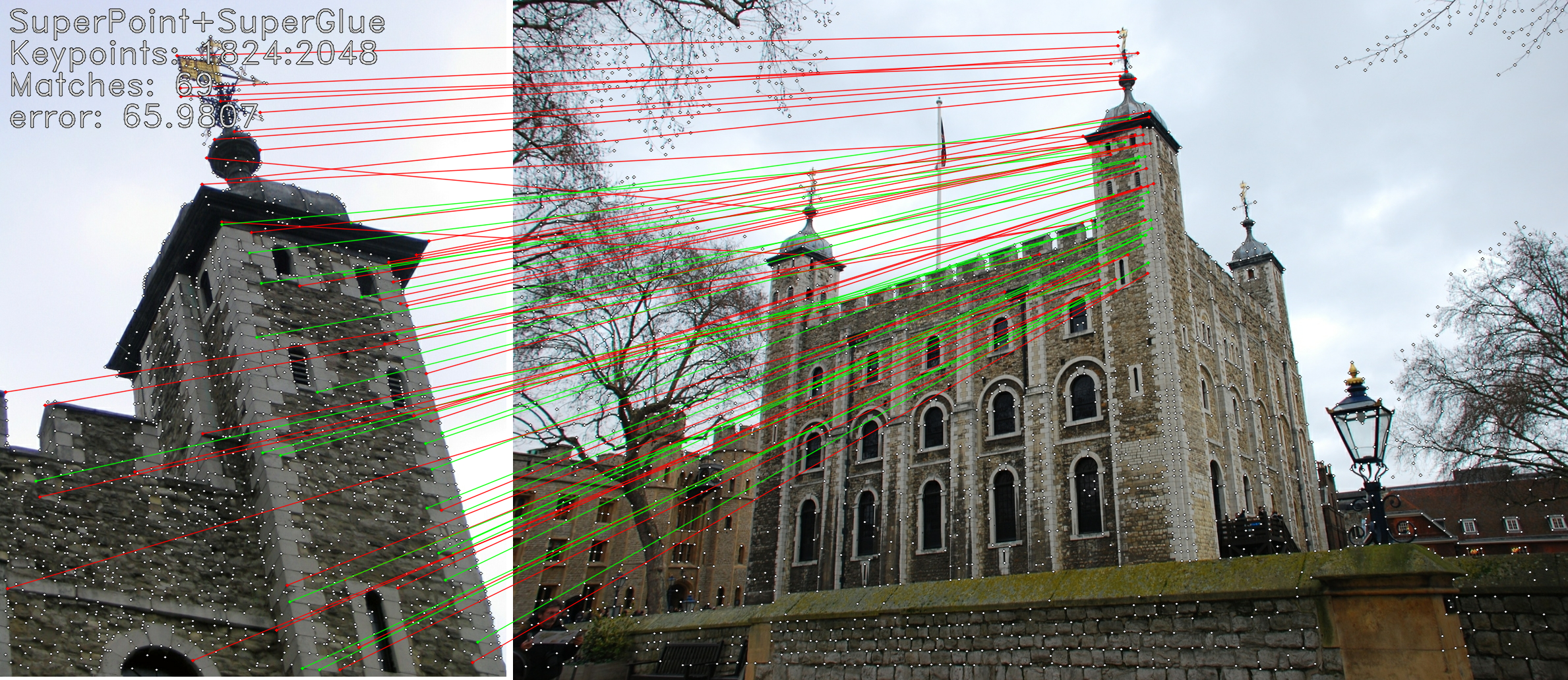}
	\end{minipage}
	\label{fig:origin}
    } \\
	\subfigure[Add OETR to guide SuperPoint and SuperGlue.]{
		\begin{minipage}[b]{0.45\textwidth}
  	 	\includegraphics[width=1\textwidth]{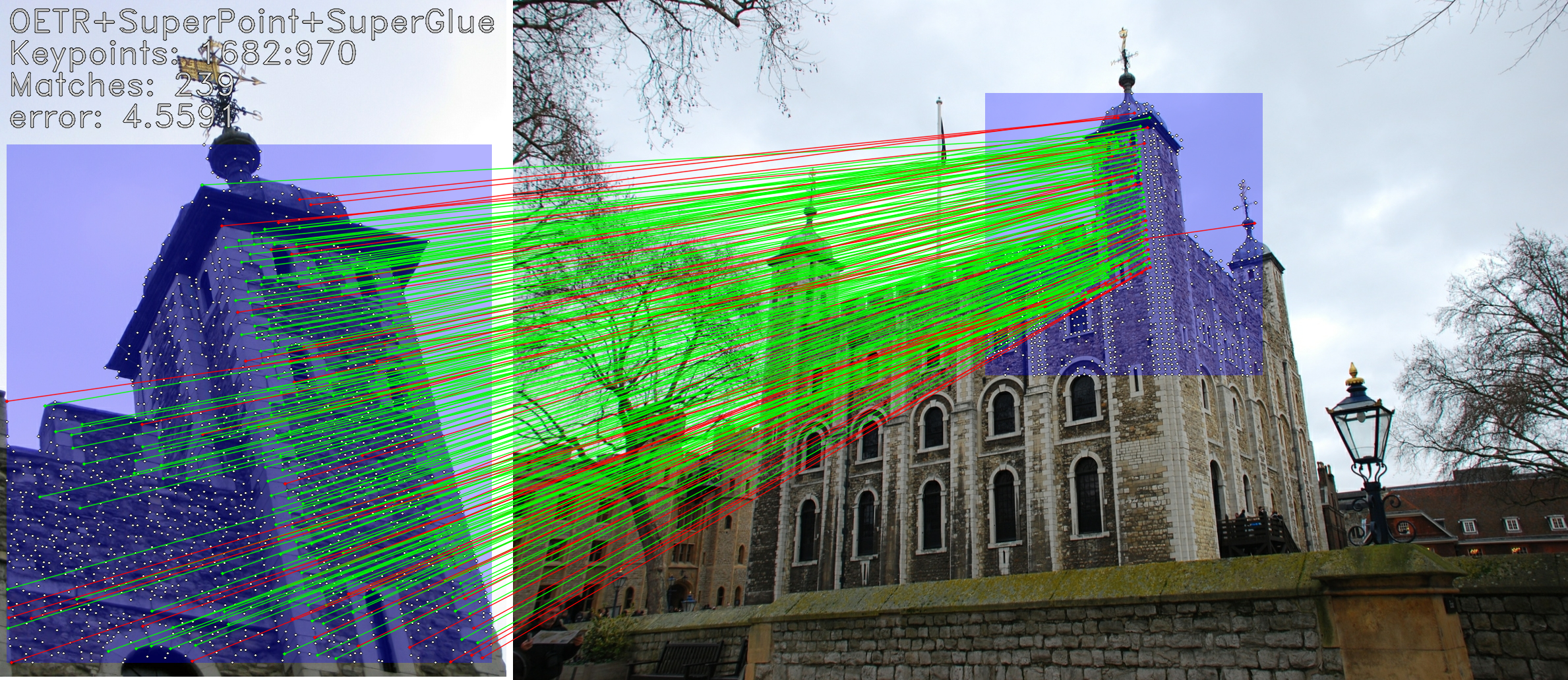}
		\end{minipage}
	\label{fig:oetr}
	}
\caption{\textbf{SP+SG vs. OETR+SP+SG}. By overlap estimation, OETR is capable of constraining local feature matching in the commonly visible regions, compensating for viewpoint change and pruning ambiguous matches.}
\label{fig1}
\end{figure}
\section{Introduction}
Detecting precise locations for local features, then establishing their reliable correspondences across images are underpinning steps towards many computer vision tasks, such as Structure-from-Motion (SfM) \cite{schonberger2016structure, wu2013towards}, visual tracking \cite{yan2021learning, Voigtlaender2020} and visual localization \cite{sarlin2019coarse}. By extension, feature matching enables real applications such as visual navigation of autonomous vehicles and portable augmented/mixed reality devices. However, under extreme appearance, viewpoint or scale changes in long-term conditions \cite{Sattler2018}, repeatable keypoints detection, and stable descriptor matching are very challenging and remain unsolved.

Traditionally, appearance, viewpoint, and scale invariance are parameterized by hand-crafted transformation and statistics of local feature patches \cite{Lowe2004, bay2006surf}. Recently, convolutional neural networks (CNNs) based local features \cite{Detone2018, revaud2019r2d2, tyszkiewicz2020disk} with strong semantic representation and attention aided matching protocols \cite{wiles2021co, sarlin2019coarse} have shown significant improvements over their hand-crafted counterparts under appearance changing conditions, such as day-night, weather, and seasonal variations. Nevertheless, detection from the deepest layer embedding high-level information often struggles to identify low-level structures (corners, edges, etc.) where keypoints are often located, leading to less accurate keypoints \cite{Germain2020}. So recent methods \cite{luo2020aslfeat} fuse earlier layers that preserve high-frequency local details to help retrieve accurate keypoints. However, corresponding descriptors are vulnerable to large view angle or scale change due to a limited receptive field that implies less semantic context. So, the performance is highly depend on complicated multi-scale feature interaction design which is not straightforward. Moreover, this dilemma becomes more severe when the commonly visible region between image pairs is limited, leading to extreme scale variations. As a result, finding stable correspondences between query and database images taken from scenes with small shared regions bottlenecks the performance of loop-closure in the context of SLAM, visual localization, or registering images to Structure-from-Motion (SfM) reconstructions. 

In this paper, we refer to a straightforward yet effective preprocessing approach to guide feature matching by estimating overlap between image pairs. Based on overlap estimation, the scale for a shared scene can be aligned prior to feature detection and description, which satisfies the scale-invariant requirement for local features starting from SIFT \cite{Lowe2004}. Meanwhile, similar to guided matching \cite{Darmon2020}, relying exclusively on local information to match images can be misleading especially in the case of scenes with repeated patterns. Our strong overlap constraint will generate disambiguated coarse prior, to prune possible outliers outside overlapped area. As shown in Fig. \ref{fig:origin}, overwhelming noisy and ambiguous feature pairs are introduced by SuperPoint detector and \cite{detone2018superpoint} SuperGlue matcher \cite{sarlin2020superglue} when the viewpoint changes. Typically, when reconstructing scenes from Internet photos, scale and viewpoint variations of the collected images will hinder stable feature matching thus degrading reconstruction performance. 

To this end, it is important to guarantee a robust and precise overlap estimation, which however is not a well-studied topic. Related areas cover few-shot object detection \cite{Fan2020}, template tracking \cite{Zhang2021, zhang2020ocean}, and most closely normalized surface overlap (NSO) presented by \cite{Rau2020}. Intuitively, estimating precise overlap bounding box between image pairs is more challenging, as it requires iterative and reciprocal validation to find shared regions from image pairs, with no initial template provided. Nevertheless, we borrow ideas from these well-studied tasks and propose a novel transformer-based correlation feature learning approach to regress precise overlap bounding boxes in image pairs.

To summarise, we make three contributions:
\begin{itemize}
\item We propose an efficient overlap estimation method to guide local feature matching, compensating for potential mismatch in scales and viewing angles. We demonstrate overlap estimation can be plugged into any local feature matching pipeline, as a preprocessing module.
\item A carefully redesigned transformer encoder-decoder framework is adopted to estimate overlap bounding boxes in image pairs, within a lightweight multi-scale feature correlation then overlap regression process. Training can be supervised by a specifically designed symmetric center consistency loss.
\item Extensive experiments and analysis demonstrate the effectiveness of the proposed method, boosting the performance of both traditional and learning-based feature matching algorithms, especially for image pairs with the small commonly visible regions.
\end{itemize}

\begin{figure*}[t]
\centering
\includegraphics[width=0.9\textwidth]{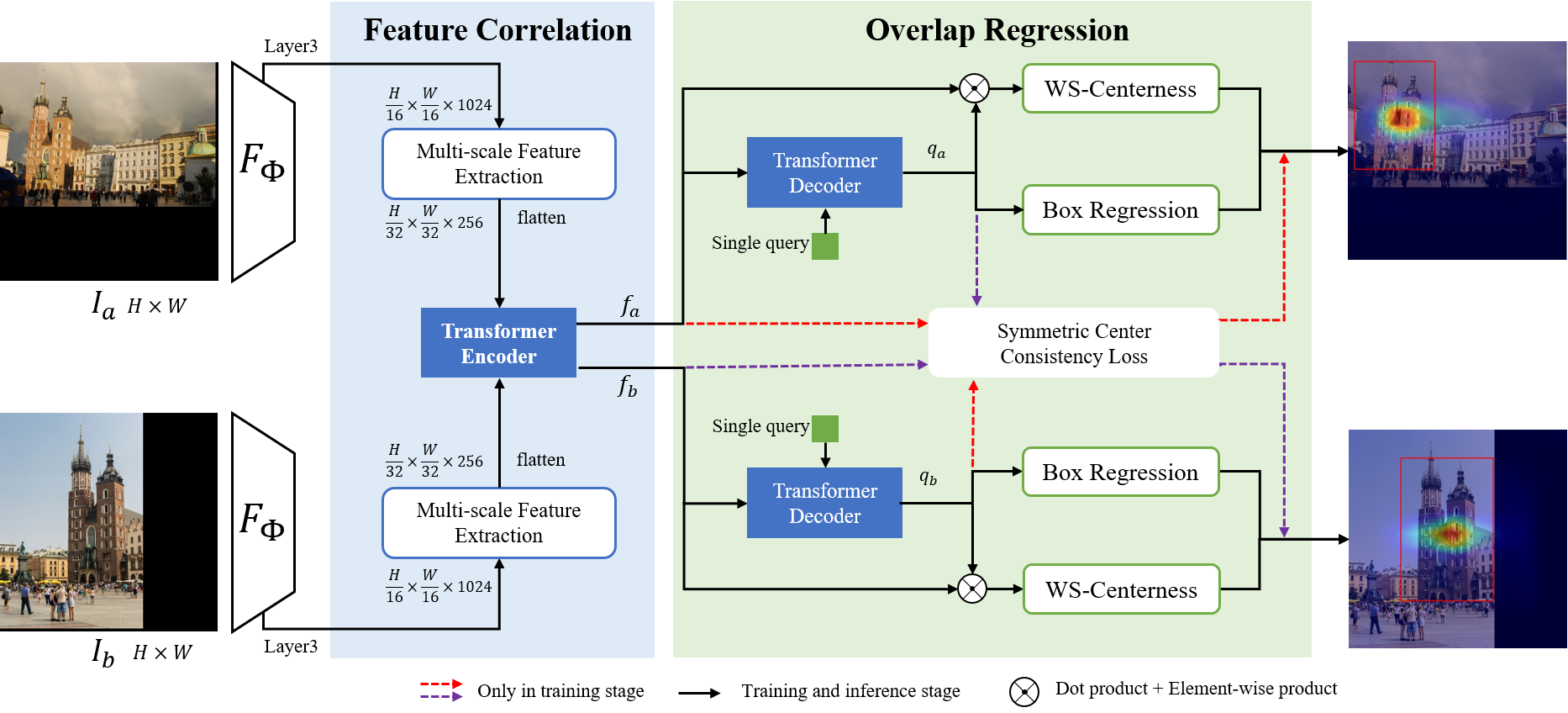}
\caption{\textbf{Overview}. OETR estimates overlap bounding boxes for image pairs with two steps: Feature Correlation and Overlap Regression. In feature correlation, with the output of backbone features, we first do convolution with three different size kernels and do self-cross attention in the Transformer encoder module. A Transformer decoder then takes a single learnable query and correlated features as inputs to regress the overlap bounding box.}
\label{overview}
\end{figure*}

\section{Related Works}
Our overlap estimation is mainly intended to guide and constrain local feature matching while regressing overlap bounding box borrows ideas from object detection.  
\subsection{Local feature matching}
SIFT \cite{Lowe2004} and ORB \cite{Rublee2011} are arguably the most renowned hand-crafted local features, facilitating many downstream computer vision tasks. Reliable local feature is achieved by hand-designed patch descriptor according to gradient-based statistics. Borrowing semantic representation ability from convolution neural networks (CNNs), robustness of local features on large appearance, scale and viewpoint change can be improved by a large margin with learning-based method \cite{Yi2016, detone2018superpoint, dusmanu2019d2, revaud2019r2d2, Luo2019, luo2020aslfeat, tyszkiewicz2020disk}. SuperGlue \cite{sarlin2020superglue} proposes a GNN based approach for local feature matching, which builds a matching matrix from two sets of keypoints with descriptors and positions. \cite{wiles2021co} proposed spatial attention mechanism for conditioning the learned features on both images under large viewpoint change. Our work is inspired by SuperGlue \cite{sarlin2020superglue} and CoAM \cite{wiles2021co} in terms of using self and cross attention in GNN for spatial-wise feature correlation. SuperGlue achieves impressive performance and sets the new state-of-the-art in local feature matching. Nevertheless, for existing local feature matching methods, our OETR can be utilized as a preprocessing module to constrain keypoint detection and descriptor matching within overlapped area. 

Besides salient keypoint detection and discriminative descriptor learning, the classical image matching pipeline performs correspondence pruning by the bidirectional ratio test \cite{Lowe2004}. More elaborate techniques such as GMS \cite{Bian2017} and LPM \cite{Ma2017} remove ambiguous matches based on the observation that matches should be consistent within close neighboring areas. Geometric verification is then performed in a RANSAC, PROSAC \cite{prosacChum2005} or recent neural guidance RANSAC \cite{ngransacBrachmann2019} based scope, to ensure epipolar geometry constraint. These methods adopt a hypothesize-and-verify approach and attempt to obtain the smallest possible outlier-free subset to estimate a provided parametric model by resampling. On the contrary, our OETR leverage overlaps constraints before matching and can help to identify correct matches. 
 
\subsection{Overlap Estimation}
\cite{Rau2020} propose a box embedding to approximate normalized surface overlap (NSO) asymmetrically. NSO is defined as the percentage of commonly visible pixels over each image, for image retrieval or pre-scale whole image accordingly for better local feature matching. By zooming in and cropping commonly visible regions around coarse matches, COTR \cite{jiang2021cotr} achieves greater matching accuracy recursively. Their overlap estimation is not straightforwardly represented by a bounding box covering a commonly visible region. Instead, we hope local feature matching can benefit more from our precise overlap bounding box estimation.

\subsection{Object Detection}
Object detection aims at localizing bounding boxes and recognizing category labels for objects of interest in one image. Mainstream one-stage detectors rely on dense positional candidates enumerating feature map grid, such as anchors boxes \cite{liu2016ssd,Lin2017,redmon2017yolo9000} and reference points \cite{tian2019fcos}, to predict final objects. As an extension, two-stage detectors \cite{Ren2015} predict foreground proposal boxes from dense candidates. Recently, sparse candidates like learnable proposals \cite{sun2021sparse} or object queries \cite{carion2020end} have been adopted to guide detection and achieved promising performance. Comparably, overlap estimation is to localize the unique bounding box of common area in each image, which is conditioned on image pairs and with no prior instance of scene information. From dense to sparse, then from sparse to unique, our overlap estimation follows objection detection to guarantee precise overlap bounding box regression.
Moreover, compared to visual object tracking (VOT) which localizes provided objects in sequential images \cite{yan2021learning}, no initial template is available for overlap estimation, thus making the spatial relationships of overlapped area more complicated \cite{Rau2020}. 


\section{Method}
In this section, we present the Overlap Estimation network with TRansformer \cite{Vaswani2017}, shorten as \textbf{OETR}. The task of overlap estimation conditioned on image pair is to predict one bounding box for each image, which tightly covers the commonly visible region as the mask shown in Fig. \ref{fig:oetr}. 

To our best knowledge, overlap estimation is not a well-studied problem. As shown in Fig. \ref{overview}, OETR estimates overlap in two steps: correlating multi-scale CNN features then regressing overlap bounding box. We call them to feature correlation neck and overlap regression head respectively, analogous to objection detection convention \cite{Ren2015}. To remedy the potential scale variance from CNN features, an efficient multi-scale kernel operator is employed. The Transformer encoder performs feature correlation by self-attention and cross-attention of flattened multi-scale features from image pair. Inspired by DETR \cite{carion2020end} and FCOS \cite{tian2019fcos}, we cast the overlap estimation problem into identifying and localizing commonly visible regions in image pairs.

\subsection{Feature Correlation}
The feature correlation step consists of a multi-scale feature extraction from CNN backbone, and a transformer feature encoder.
\subsubsection{Multi-scale Feature Extraction}
\begin{figure}[t] 
\centering
\includegraphics[width=0.45\textwidth]{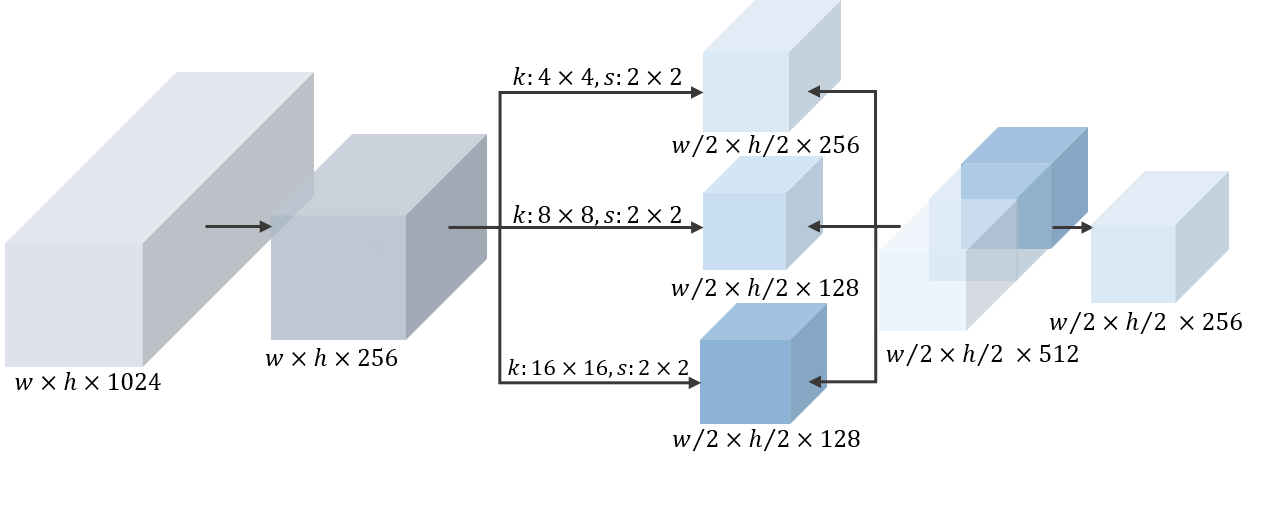} 
\caption{Our design choice for multi-scale feature extractor: shared layer3 from ResNet50 is convoluted by three kernels (i.e., $4\times4$, $8\times8$, $16\times16$) with stride $2\times2$, then concatenated in channel dimension.}
\label{msf}
\end{figure}

Commonly used methods for multi-scale feature extraction are feature pyramid network (FPN) \cite{lin2017feature} and its variants \cite{Liu2018} \cite{kirillov2019panoptic}, which output proportional size feature maps at multiple levels by different convolutional strides. However, feature correlation between multiple levels' feature map is computationally intensive. Assuming correlating 4 layers (P2, P3, P4, P5) of FPN, 16 times cross feature map correlation are required. To this end, we adopt a lightweight Multi-Scale kernel Feature extractor (MSF) \cite{Wang2021}, as shown in  Fig.\ref{msf}. MSF first employs three kernel operators in parallel on layer3 from ResNet50, with stride of 2. Three convoluted features are then concatenated in channel dimension, blending the output embedding with multi-scale feature patches whose receptive fields are more flexible. Meanwhile, we leverage a lower channel dimension for large kernels while a higher dimension for small kernels, to balance computational cost.

\subsubsection{Transformer encoder}
Considering that overlapped area shares common scene information between image pairs, final overlap bounding box in each image is conditioned on features from its own and paired image. To facilitate efficient feature interaction between image pairs, We inherit the core design of popular iterative self-attention and cross-attention \cite{sarlin2020superglue, JiamingSunHujunBao2021} and propose a lightweight linear transformer\cite{katharopoulos2020transformers} encoder layer for message passing within and across image pairs. 

\begin{figure}[t] 
\centering
\includegraphics[width=0.4\textwidth]{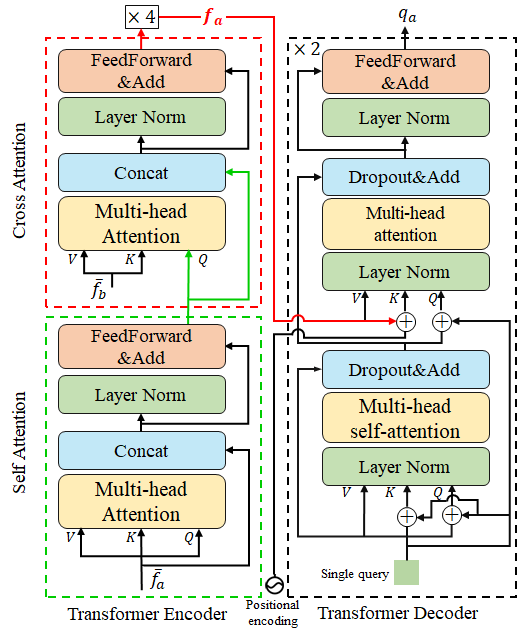} 
\caption{\textbf{Redesigned transformer encoder and decoder architecture for overlap estimation.} Feature correlation is achieved by $4\times$ self-cross attention layers with flattened $\bar{f}_{a}$ and $\bar{f}_{b}$ as input. Combined with single query, correlated feature $f_{a}$ is then fed into transformer decoder to obtain $q_{a}$.}
\label{transformer}
\end{figure}

Different from template matching methods \cite{Fan2020, Zhang2021}, image $I_a$ is not always part of image $I_b$ for overlap estimation problem. To embed variant spatial relationships of overlapped area from paired image with unpredictable scale, viewpoint or appearance changes, we directly flatten the multi-scale features from MSF, then complete the feature correlation by transformer encoder. Adapted from vanilla Transformer \cite{Vaswani2017} with only self-attention layer, our Transformer encoder correlates features from paired image by iterative self-attention and cross-attention layers which are identical to that used by \cite{sarlin2020superglue, JiamingSunHujunBao2021}. The detail components of Transformer encoder are presented in left side of Fig. \ref{transformer}. For multi-scale flattened feature $ \bar{f}_a$ from image ${I_a}$, self-attention is focused on internal correlation $\bar{f}_a$, then cross-attention correlates features from $ \bar{f}_b$. This message-passing operator is interleaved by 4 times, ensuring sufficient feature interaction between image pair. In order to make better use of the relative position relationship in spatial. Different from LoFTR\cite{JiamingSunHujunBao2021}, we add positional encoding to $ \bar{f}_a$ and $ \bar{f}_b$ in every iteration. 


\subsection{Overlap Regression}
For overlap estimation, only one bounding box covering a commonly visible region should be regressed. We borrow the idea from DETR \cite{detrCarion}, which learns different spatial specializations for each object query, performs co-attention between object queries and encoded features with the Hungarian algorithm \cite{kuhn1955hungarian} for prediction association. To guarantee unique overlap prediction, single learnable query is employed to reason its relation to the global image context. After feature correlation, $f_{a}$ and $f_{b}$ are fed into a transformer decoder with single query. The detail components of transformer decoder is illustrated in the right side of Fig. \ref{transformer}. 

Overlap regression can be regarded as surrogate two sub-problems: overlapped area center localization and bounding box side offset regression, which is inspired by FCOS \cite{tian2019fcos}. FCOS introduces a lightweight center-ness branch to depict the distance of a location to the center of its corresponding bounding box, and a regression branch to predict the offsets from the center to four sides of the bounding box. The proposed overlap regression inherits FCOS's design and takes decoded feature $q_{a}$ (or $q_{b}$) and correlated feature $f_{a}$ (or $f_{b}$) as inputs, as shown in the right side of Fig. \ref{overview}. For WS-Centerness branch, the similarity between correlated feature $f_{a}$ and the decoded feature $q_{a}$ can be computed by dot-product operation. Next, the similarity scores are element-wisely multiplied with correlated features, to enhance attention on the overlapped areas while weakening attention on the non-overlapped areas. 

The generated feature vector is reshaped to a feature map and fed into a fully convolutional network (FCN), generating center coordinate probability distribution $P_c(x, y)$. True centerness of the overlapped area is then obtained by computing the expectation of the center coordinate's probability distribution as shown in Eq. \ref{center}, which is weighted-sum (WS) of center coordinate by center probability.

For the box regression branch, only decoder feature $q_{a}$ is utilized to regress a 4-dimensional vector $(l,t,r,b)$, which is the offset from the overlapped area center to four sides of the bounding box. Final overlap bounding box is localized by the center location and predicted $(l,t,r,b)$.

\begin{equation}
\left(\widehat{x}_{c}, \widehat{y}_{c}\right)=\left(\sum_{y=0}^{H} \sum_{x=0}^{W} x \cdot P_{c}(x, y), \sum_{y=0}^{H} \sum_{x=0}^{W} y \cdot P_{c}(x, y)\right)
\label{center}
\end{equation}

\subsection{Symmetric Center Consistency Loss}
Consistency loss is commonly employed in feature matching pipelines \cite{wang2019learning}. For overlap estimation, we hope a single query for each image should be close in feature space, as they represent the commonly visible regions. However, due to potential large appearance or viewpoint changes, sharing a common query for paired images is not sufficient. To provide consistency supervision, we introduce symmetric center consistency loss, which ensures forward-backward mapping of the overlapped area center to be spatially close. Given image pair $I_a$ and $I_b$, the output ($f_a$, $f_b$) of feature correlation is embedded with decoder output ($q_a$, $q_b$) as shown in Fig. \ref{overview}. We also embed ($q_a$, $q_b$) to ($f_b$, $f_a$) respectively, for center consistency. Finally, same as DETR \cite{detrCarion}, L1 loss, and generalized IoU loss are introduced for box localization.

\begin{equation}
\begin{aligned}
\mathcal{L}=\sum_{i=a}^b(\lambda_{con}\left\|c_{i}-\widetilde{c}_{i}\right\|_{1}+\lambda_{loc}\left\|c_{i}-\hat{c}_{i}\right\|_{1} \\+\lambda_{iou}\mathcal{L}_{iou}(b_{i}, \hat{b}_{i})+\lambda_{L1}\left\|b_{i}-\hat{b}_{i}\right\|_{1})
\end{aligned}
\label{eq4}
\end{equation}

where $c_i, \hat{c_i}$ and $\widetilde{c_i}$ represent the groundtruth, prediction and symmetric consistency center position of overlap bounding box, respectively. Note that center position here refers to geometric center of bounding box, different with $(\widehat{x}_{c},\widehat{y}_{c})$ in Eq. \ref{center}. $b_{i}\in [0, 1]^{4}$ is a vector that defines groundtruth box center coordinates and its height and width relative to the image size. $b_i$ and $\hat{b_i}$ represent the groundtruth and the predicted box respectively.  $\lambda_{con}, \lambda_{loc}, \lambda_{iou}$ and $\lambda_{L1} \in \mathbb{R}$ are hyper-parameters to balance losses.

\section{Experiments}

\subsection{Implementation Details}

\subsubsection{Training.} We train our overlap estimation model \textbf{OETR} on MegaDepth \cite{li2018megadepth} dataset. Image pairs are randomly sampled offline, with overlap ratio in [0.1, 0.7]. According to IMC2021 \cite{jin2021image} evaluation requirements, we remove overlapping scenes with IMC's validation and test set from MegaDepth. Overlap bounding box groundtruth is calculated from provided depth, relative pose and intrinsics of image pairs. To enable batched training, input images are resized to have their longer side being 1200 while image ratio is kept, followed by padding to 1216 (can be divided by 32) for both sides. The loss weights $\lambda_{con}, \lambda_{loc}, \lambda_{iou}$ and $\lambda_{L1}$ are set to $[1, 1, 0.5, 0.5]$ respectively. The model is trained using AdamW with weight decay of $10^{-4}$ and a batch size of 8. It converges after 48 hours of training on 2 NVIDIA-V100 GPUs with 35 epochs.

\subsubsection{Inference.} 
\begin{figure}[H] 
\centering
\includegraphics[width=0.45\textwidth]{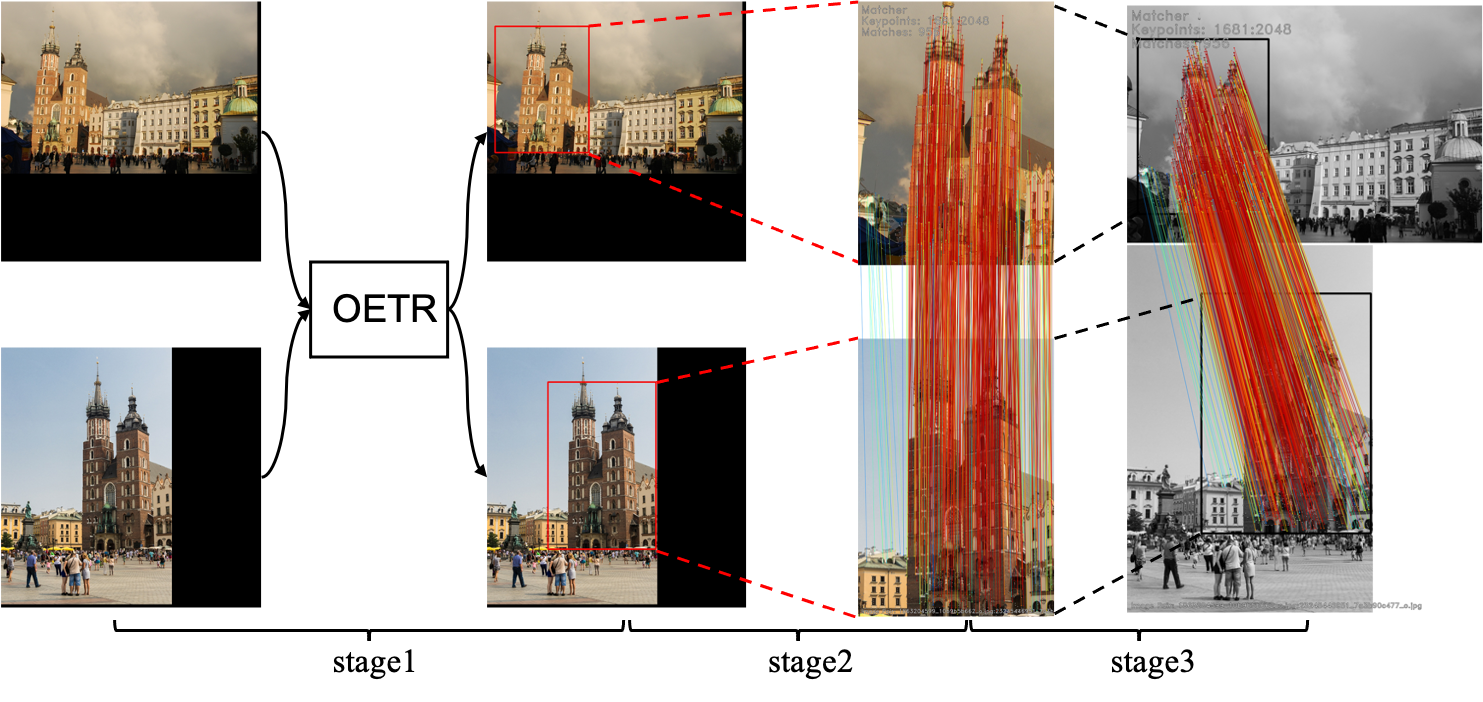} 
\caption{\textbf{OETR as the preprocessing module for local feature matching.}}
\label{inference}
\end{figure}

In this section, we discuss how to apply OETR as the preprocessing module for local feature matching. As shown in Fig.
\ref{inference}, there are three stages: 1) Resized and padded image pair ($1216\times 1216$) is fed into OETR for overlap estimation. 2) Overlapped areas are cropped out and resized to mitigate potential scale mismatch. The resized ratio is the product of the origin image resize ratio and overlap scale ratio. The overlap scale ratio is calculated by:
\begin{equation}
\begin{aligned}
s(O_A, O_B)=max(\frac{w_A}{w_B}, \frac{w_B}{w_A}, \frac{h_B}{h_A}, \frac{h_A}{h_B})
\end{aligned}
\label{scale}
\end{equation}
where $O_A$ and ${O_B}$ are overlapping bounding boxes for image pair $A$ and $B$, with their width and height as ${(w_A, h_A), (w_B, h_B)}$ respectively.
3) Local feature matching is performed on cropped overlap aligned images. Finally, we warp keypoints and matches back to origin images and perform downstream tasks such as relative pose estimation.

\subsection{Comparison with Existing Methods}
We add our OETR as a preprocessing module with different feature extractors (SuperPoint \cite{Detone2018}, D2-Net \cite{dusmanu2019d2}, Disk \cite{tyszkiewicz2020disk}, R2D2 \cite{revaud2019r2d2} and matchers (SuperGlue \cite{sarlin2020superglue}, NN), and evaluate it on two benchmarks: MegaDepth \cite{li2018megadepth} and IMC2021 \cite{jin2021image}.

\begin{figure*}[t] 
\centering
\includegraphics[width=1.0\textwidth]{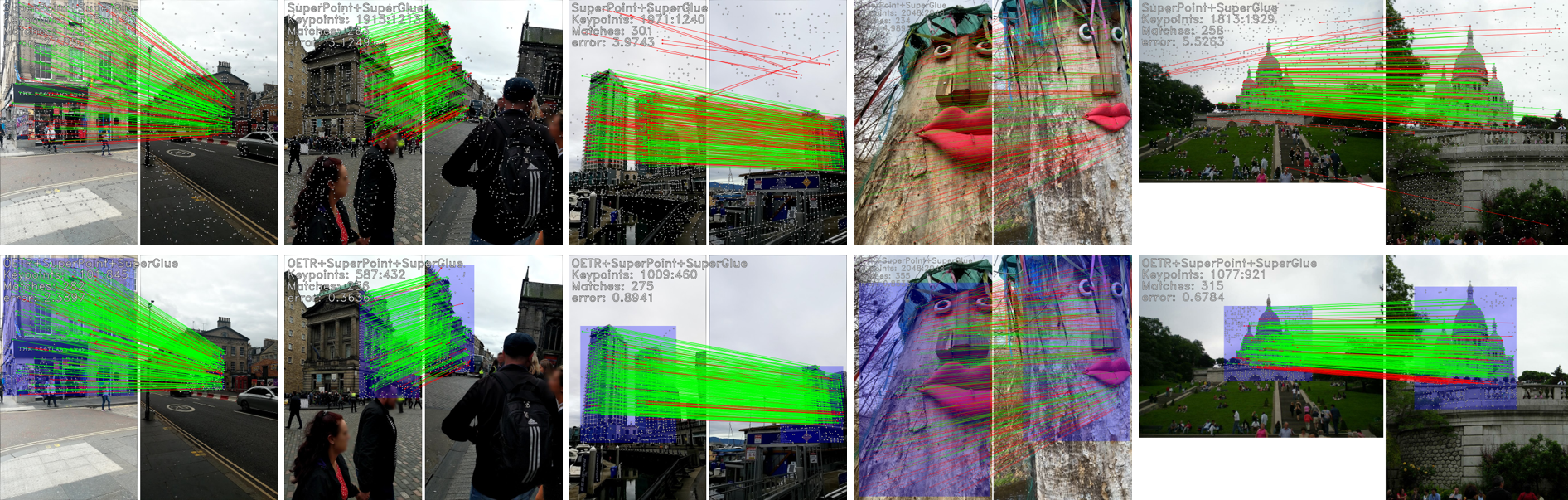} 
\caption{\textbf{Visualizing MegaDepth matching results.}. Adding OETR can consistently generate more correct matches (green lines) and fewer wrong matches (red lines), especially for image pairs with the small overlapped areas.}
\label{IMC}
\end{figure*}

\begin{table*}[t]
\centering
\resizebox{\textwidth}{!}{
\begin{tabular}{@{}ccccccccccccccccccc@{}}
\toprule\midrule&\multicolumn{6}{c}{\textbf{GoogleUrban}} &\multicolumn{6}{c}{\textbf{PragueParks}} &\multicolumn{6}{c}{\textbf{Phototourism}} \\
\cmidrule(l){2-19}&\multicolumn{3}{c}{\textbf{AUC}} &&&&
\multicolumn{3}{c}{\textbf{AUC}} &&&& \multicolumn{3}{c}{\textbf{AUC}} &&&\\ \cmidrule(lr){2-4} \cmidrule(lr){8-10} \cmidrule(lr){14-16}
\multirow{-3}{*}{\textbf{Methods}} &@$5^{\circ}$ &@$10^{\circ}$ &@$20^{\circ}$ &
\multirow{-2}{*}{\textbf{P}} &
\multirow{-2}{*}{\textbf{MS}} &
\multirow{-2}{*}{\textbf{mAA}} &@$5^{\circ}$ &@$10^{\circ}$ &@$20^{\circ}$ &
\multirow{-2}{*}{\textbf{P}} &
\multirow{-2}{*}{\textbf{MS}} &
\multirow{-2}{*}{\textbf{mAA}} &@$5^{\circ}$ &@$10^{\circ}$ &@$20^{\circ}$ &
\multirow{-2}{*}{\textbf{P}} &
\multirow{-2}{*}{\textbf{MS}} &
\multirow{-2}{*}{\textbf{mAA}} \\ 
\midrule
\textbf{D2-Net+NN} &2.34 &5.06 &9.96 &54.13 &2.35 &5.52 &27.67 &42.29 &54.58 &48.6 &2.06 &45.45 &11.79 &20.6 &31.01 &58.9 &2.77 &22.36 \\
\textbf{+OETR} &\underline{3.00} &\underline{6.89} &\underline{13.29} &\underline{59.24} &\underline{3.37} &\underline{7.591} &\underline{32.17} &\underline{47.90} &\underline{59.92} &\underline{47.74} &\underline{2.41} &\underline{51.29} &\underline{23.26} &\underline{36.87} &\underline{50.69} &\underline{68.18} &\underline{6.61} &\underline{39.75} \\ 
\midrule
\textbf{DISK+NN} &7.76 &14.62 &23.99 &76.50 &5.27 &15.93 &35.20 &52.98 &65.74 &43.48 &4.43 &56.697 &33.07 &49.32 &64.03 &83.71 &13.13 & 52.94\\
\textbf{+OETR} &\underline{9.70} &\underline{18.04} &\underline{28.82} &\underline{74.28} &\underline{7.24} &\underline{19.71} &\underline{36.89} &\underline{56.53} &\underline{69.61} &\underline{44.55} &\underline{4.85} &\underline{60.47} &\underline{47.37} &\underline{64.41} &\underline{77.38} &\underline{86.18} &\underline{17.39} &\underline{68.70} \\
\midrule
\textbf{SP+NN} &9.28 &16.85 &26.63 &69.38 &6.33 &18.31 &50.12 &68.35 &80.30 &\underline{49.65} &5.32 &72.67 &28.63 &42.96 &56.39 &68.15 &7.87 &46.12 \\
\textbf{+OETR} &\underline{9.35} &\underline{17.88} &\underline{28.92} &\underline{70.90} &\underline{9.33} &\underline{19.50} &\underline{53.89} &\underline{72.66} &\underline{84.48} &48.40 &\underline{7.61} &\underline{77.30} &\underline{41.12} &\underline{57.89} &\underline{71.98} &\underline{74.74} &\underline{15.88} &\underline{61.90} \\
\midrule
\textbf{R2D2(MS)+NN} &12.96 &24.54 &38.69 &66.70 &4.15 &26.62 &\underline{55.14} &\underline{75.15} &\underline{86.93} &\underline{47.85} &7.42 &\underline{80.10} &43.39 &61.88 &76.56 &74.07 &7.02 &66.22 \\
\textbf{+OETR} &
\underline{14.91} &\underline{26.23} &\underline{39.94} &\underline{67.14} &\underline{5.91} &\underline{28.47} & 54.04 &73.32 &84.99 &47.23 &\underline{9.08} &78.00 &\underline{53.49} &\underline{70.47} &\underline{82.62} &\underline{80.18} &\underline{15.83} &\underline{74.95}\\\midrule
\textbf{SP+SG} &
15.60 &27.46 &41.82 &73.64 &13.38 &29.71 &\underline{61.39} &\underline{79.07} &\underline{89.21} &\textbf{50.97} &11.05 &\underline{84.02} &48.86 &67.10 &80.97 &74.47 &17.56 &71.64 \\
\textbf{+OETR} &
\underline{16.82} &\underline{29.56} &\underline{44.26} &\underline{75.50} &\underline{19.36} &\underline{32.09} &60.14 &78.43 &88.71 &49.35 &\textbf{14.06} &83.46 &\underline{55.74} &\underline{72.19} &\underline{84.02} &\underline{79.59} &\underline{29.50} &\underline{76.66} \\ \midrule
\textbf{DISK+SG} &
17.25 &30.19 &45.53 &73.80 &14.14 &32.74 &51.70 &71.82 & 84.54 &\underline{47.83} &11.24 &76.58 &52.23 &70.09 & 83.17 &80.90 &32.25 &74.64 \\
\textbf{+OETR} &
\underline{19.77} &\underline{32.67} &\underline{47.17} &\underline{74.28} &\underline{19.64} &\underline{35.35} &\underline{52.43} &\underline{72.18} &\underline{84.57} &47.80 &\underline{11.29} &\underline{76.93} &\textbf{59.91} &\textbf{75.53} &\textbf{86.16} &\textbf{83.51} &\textbf{38.18} &\underline{79.99} \\ \midrule
\textbf{SP+SG*} &
18.21 &31.74 &47.15 &76.68 &14.99 &34.35 &64.36 &\textbf{81.36} &\textbf{90.49} &50.60 &10.36 &\textbf{86.27} &52.65 &70.43 &83.31 &77.82 &18.74 &75.04 \\
\textbf{+OETR} &
\textbf{19.28} &\textbf{32.99} &\textbf{48.57} &\textbf{77.61} &\textbf{20.79} &\textbf{35.80} &\textbf{64.72} &81.12 &90.33 &\underline{50.62} &\underline{10.46} &86.15 &\underline{59.75} &\underline{75.46} &\underline{86.08} &\underline{82.37} &\underline{31.00} &\textbf{80.01} \\ \midrule
\bottomrule
\end{tabular}}
\caption{\textbf{Stereo performance on IMC2021.} We report AUC at $5^{\circ}$, $10^{\circ}$ and $20^{\circ}$, matching precision, matching score, and mean Average Accuracy (mAA) at $10^{\circ}$, similarly as official leaderboard evaluation protocol. With identical local feature extractor and matcher, we highlight better method in \underline{underline} when compared with adding OETR as the preprocessing module. We further highlight best method overall in \textbf{bold}.}
\label{imctable}
\end{table*}

\subsubsection{Metrics}
Following \cite{sarlin2020superglue}, we report the \textbf{AUC} of the pose error under thresholds ($5^{\circ}, 10^{\circ}, 20^{\circ}$), where the pose error is set as the maximum angular error of relative rotation and translation. Following IMC2021 \cite{jin2021image}, we additionally use \textbf{mAA} (mean Average Accuracy) up to a 10-degree error threshold. In our evaluation protocol, the relative poses are recovered from the essential matrix, estimated from feature matching with RANSAC. We also report match precision(\textbf{P}) and matching score(\textbf{MS}) in normalized camera coordinates, with epipolar distance threshold of $5\
\cdot 10^{-4}$ \cite{Detone2018, dusmanu2019d2, sarlin2020superglue} .

\subsubsection{IMC2021}
IMC2021 is a benchmark dataset for local feature matching competition, whose goal is to encourage and highlight novel methods for image matching that deviate from and advance traditional formulations, with a focus on large-scale, wide-baseline matching for 3D reconstruction or pose estimation \cite{jin2021image}. There are three leaderboards: \textit{Phototourism}, \textit{PragueParks}, and \textit{GoogleUrban}. They focus on different scenes but all measure the performance of real problems. The challenge features two tracks: stereo, and multi-view (SfM) and we focus on the stereo task. We summarize the results of IMC2021 validation datasets in Tab.\ref{imctable}. Noted that the official training code of SuperGlue is not available and its public model (denoted as SG*) is trained on full MegaDepth dataset which has overlapping scenes with \textit{Phototourism}. Instead, we retrain SuperGlue with different extractors (SuperPoint and DISK) on MegaDepth without the pretrained model and remove scenes sharing with IMC2021's validation and test set.


As shown in Tab. \ref{imctable}, in Phototourism and GoogleUrban, matching performance is improved for all existing methods after adding OETR. However, in PragueParks, we observe a slight performance degradation for SP+SG(SG*) and R2D2(MS)+NN. Moreover, we claim that this is mainly due to unnoticeable scale differences in PragueParks, thus slightly inaccurate overlap bounding box estimation would prune correct matches, especially those near overlap border. For feature matching like SP+SG(SG*) or multi-scale R2D2 which show strong matching ability, performance can hardly be influenced by adding OETR for image pairs with nearly identical scales. This assumption can be further proved by following experiments on the scale-separated MegaDepth dataset.

\subsubsection{MegaDepth}
\begin{figure}[]
\centering
\includegraphics[width=0.45\textwidth]{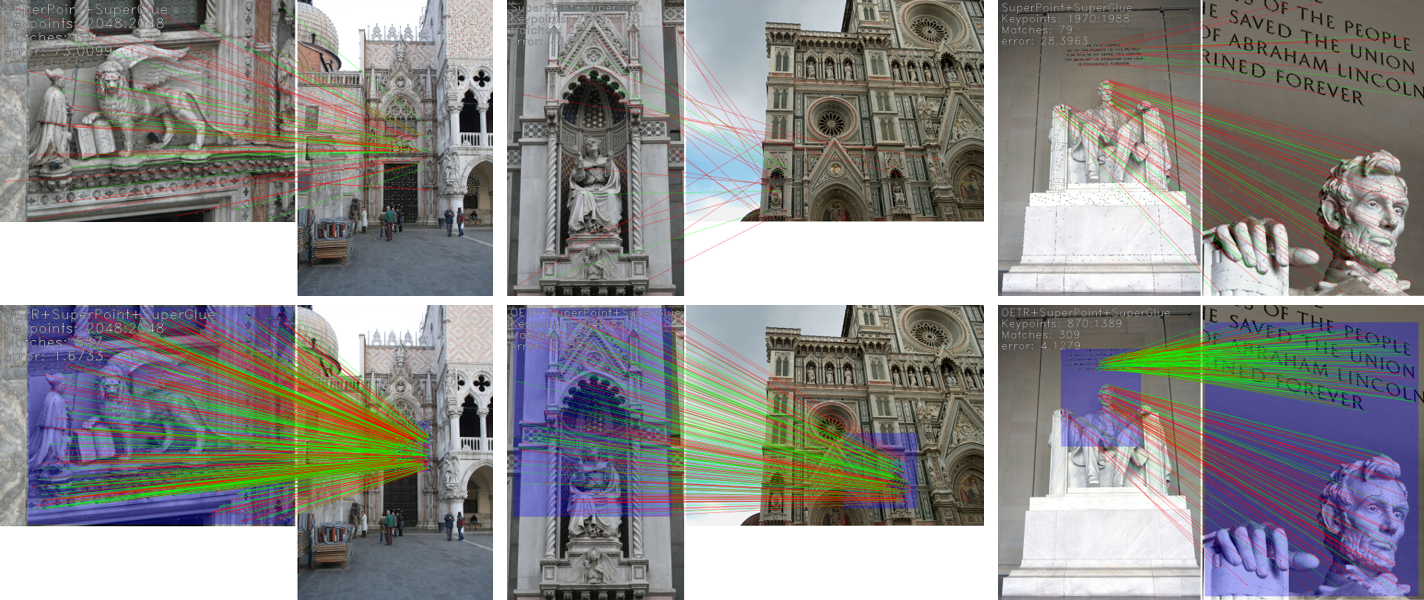} 
\caption{\textbf{Visualizing MegaDepth matching results.} Original SP+SG tends to generate matches deviate from epipolar constrain. Adding OETR substantially improve matching and thus pose estimation performance.}
\label{fig7}
\end{figure}

We split MegaDepth test set (with 10 scenes) into subsets according to the overlap scale ratio as in Eq. \ref{scale} for image pairs. We separate overlap scales into $[1, 2), [2, 3), [3, 4), [4, +\infty)$ and combine $[2, 3), [3, 4), [4, +\infty)$ as $[2, +\infty)$ for image pairs with noticeable scale difference. Fig. \ref{fig7} qualitatively shows the comparison when adding OETR before image matching.

We first compare the results of different feature extraction and matching algorithms on MegaDepth $[2, +\infty)$ before and after adding OETR as the preprocessing module. OETR consistently outperforms the plain method as shown in Tab.\ref{tab1}, especially for NN matching. For strong matching baseline SuperGlue, we also observe a noticeable performance improvement. As shown in Tab.\ref{tab2}, the larger the scale variation between image pairs, the more obvious performance gain will be obtained by adding OETR. Artificially aligning the commonly visible region to a nearly identical scales can alleviate potential viewpoint mismatch. SG and SG* indicates our own trained model and open-sourced model respectively.

\begin{table}[t]
\centering
\resizebox{.42\textwidth}{!}{
\begin{tabular}{@{}cccccc@{}}
\toprule
\multirow{2}{*}{\textbf{Methods}} & \multicolumn{3}{c}{\textbf{AUC}} & \multirow{2}{*}{\textbf{P}} & \multirow{2}{*}{\textbf{MS}} \\ \cmidrule(lr){2-4} &@$5^{\circ}$ &@$10^{\circ}$ &@$20^{\circ}$ &&\\ 
\midrule
\textbf{DISK+NN} &1.92 &3.01 &4.22 &40.45 &0.22\\
\textbf{+OETR} &\underline{10.96} &\underline{17.16} &\underline{23.88} &\underline{54.91} &\underline{3.14}\\
\midrule
\textbf{SP+NN} &2.10 &3.63 &5.70 &54.02 &1.08\\
\textbf{+OETR} &\underline{14.21} &\underline{23.43} &\underline{33.29} &\underline{69.14} & \underline{6.08}\\
\midrule
\textbf{R2D2(MS)+NN} &12.59 &22.16 &32.96 &66.78 &2.97\\
\textbf{+OETR} &\underline{27.53} &\underline{42.51} &\underline{57.42} &\underline{80.01} &\underline{11.55}\\
\midrule
\textbf{DISK+SG} &16.03 &26.07 &37.14 &72.49 &8.42\\
\textbf{+OETR} &\underline{21.27} &\underline{33.66} &\underline{46.75} &\underline{79.05} &\underline{17.22}\\
\midrule
\textbf{SP+SG*} &24.61 &38.67 &53.49 &82.40 &11.53\\
\textbf{+OETR} &\textbf{30.07} &\textbf{46.49} &\textbf{62.45} &\textbf{87.15} &\textbf{25.39 }\\ \bottomrule
\end{tabular}}
\caption{\textbf{Evaluation on MegaDepth.} OETR consistently boosts performance for variant local features.}
\label{tab1}
\end{table}

\begin{table}[t]
\centering
\resizebox{.42\textwidth}{!}{
\begin{tabular}{@{}ccccccc@{}}
\toprule
\multirow{2}{*}{\textbf{Methods}} & \multirow{2}{*}{\textbf{Scales}} & \multicolumn{3}{c}{\textbf{AUC}}                 & \multirow{2}{*}{\textbf{P}} & \multirow{2}{*}{\textbf{MS}} \\ \cmidrule(lr){3-5}
                                  &                                  & @$5^{\circ}$ &@$10^{\circ}$ &@$20^{\circ}$             &                             &                              \\ \midrule
\textbf{SP+SG*}                            & \multirow{2}{*}{{[}1,2)}         & \textbf{50.09} & 67.12          & 79.59          & 88.27                       & 28.75                        \\
\textbf{+OETR}                       &                                  & 49.76          & \textbf{67.42} & \textbf{80.02} & \textbf{89.80}               & \textbf{41.16}               \\ \midrule
\textbf{SP+SG*}                            & \multirow{2}{*}{{[}2,3)}         & 41.55          & 58.90           & 73.36          & 85.31                       & 17.42                        \\
\textbf{+OETR}                       &                                  & \textbf{42.51} & \textbf{60.28} & \textbf{74.97} & \textbf{88.30}               & \textbf{33.30}                \\ \midrule
\textbf{SP+SG*}                            & \multirow{2}{*}{{[}3,4{)}}       & 21.07          & 36.05          & 53.12          & 83.37                       & 10.58                        \\
\textbf{+OETR}                       &                                  & \textbf{27.06} & \textbf{44.63} & \textbf{61.47} & \textbf{87.33}              & \textbf{26.57}               \\ \midrule
\textbf{SP+SG*}                            & \multirow{2}{*}{$[4, +\infty)$} & 11.30           & 21.17          & 34.09          & 78.54                       & 6.60                          \\
\textbf{+OETR}                       &                                  & \textbf{20.43} & \textbf{34.72} & \textbf{49.89} & \textbf{84.96}              & \textbf{19.09}               \\ 
\bottomrule
\end{tabular}}
\caption{\textbf{Evaluation on MegaDepth.} Performance gain from OETR becomes more prominent when scale variation between image pairs increases.}
\label{tab2}
\end{table}


\subsection{Ablation Study}
In this section, we conduct ablation study to demonstrate the effectiveness of our design choice for OETR. We evaluate five different variants with results on MegaDepth $[2, +\infty)$ subset, as shown in Tab. \ref{ablation}: 1) Substituting FCOS head (select locations fall into overlap bounding box as positive samples) for overlap regression results in a significant drop in AUC. 2) Removing the multi-scale feature extraction module results in a degraded pose estimation accuracy as expected. 3) Using the original FCOS center-ness branch as argmax indexing for a central location without weighted sum operation also leads to declined results. 4) Adding overlap consistency loss during training improves the performance. 

\begin{table}[H]
\centering
\resizebox{.45\textwidth}{!}{
\begin{tabular}{@{}lccc@{}}
\toprule
\multicolumn{1}{c}{} & \multicolumn{3}{c}{\textbf{AUC}}\\ \cmidrule(l){2-4} \multicolumn{1}{c}{\multirow{-2}{*}{\textbf{Method}}} & @$5^{\circ}$  & @$10^{\circ}$ & @$20^{\circ}$ \\ 
\midrule
1) replace head with FCOS &28.39 &44.24 &59.99\\
2) remove multi-scale feature extraction &28.84	&44.23 &59.58\\
3) remove weighted sum from WS-Centerness &27.51 &43.52	&59.12\\
4) remove overlap consistency loss &29.06 &45.79 &62.22\\
\midrule
\multicolumn{1}{c}{\textbf{OETR+SP+SG*}} & \textbf{30.07} &\textbf{46.49}	&\textbf{62.45} \\
\bottomrule
\end{tabular}}
\caption{\textbf{Ablation study.} Five variants of OETR are trained and evaluated both on the MegaDepth dataset, which validates our design choice.}
\label{ablation}
\end{table}


\section{Conclusions}
This paper introduces a novel overlap estimation architecture OETR, with redesigned transformer encoder-decoder. As a preprocessing module, OETR constrains features within the overlapped areas so that ambiguous matches outside can be pruned. Crucially, benefiting from efficient multi-scale feature correlation, OETR mitigates possible scale variations between image pairs. Our experiments show that simply plugged into existing local features matching pipeline OETR boosts their performances substantially, especially for image pairs with the small commonly visible regions. We believe that OETR introduces a new perspective to guide local feature matching. Moreover, the proposed overlap estimation problem may be a promising research direction for potential applications other than local feature matching.
\bibliography{ref.bib}
\end{document}